\documentclass{article}

    \usepackage[nonatbib,final]{neurips2019}

\usepackage[utf8]{inputenc} 
\usepackage[T1]{fontenc}    
\usepackage{hyperref}       
\usepackage{url}            
\usepackage{booktabs}       
\usepackage{amsfonts}       
\usepackage{nicefrac}       
\usepackage{microtype}      
\usepackage{graphicx}
\usepackage{amsmath}
\usepackage{subcaption}

\title{Semi-Supervised Histology Classification using \\Deep Multiple Instance Learning and \\Contrastive Predictive Coding}

\author{%
  Ming Y. Lu, Richard J. Chen, Jingwen Wang, Debora Dillon and Faisal Mahmood\\
  Division of Computational Pathology, Brigham and Woman's Hospital,\\ Harvard Medical School, Boston, MA 02115 \\
  \texttt{mlu16@bwh.harvard.edu, faisalmahmood@bwh.harvard.edu} \\
}

\begin{document}

\maketitle

\begin{abstract}
Convolutional neural networks can be trained to perform histology slide classification using weak annotations with multiple instance learning (MIL). However, given the paucity of labeled histology data, direct application of MIL can easily suffer from overfitting and the network is unable to learn rich feature representations due to the weak supervisory signal. We propose to overcome such limitations with a two-stage semi-supervised approach that combines the power of data-efficient self-supervised feature learning via contrastive predictive coding (CPC) and the interpretability and flexibility of regularized attention-based MIL. We apply our two-stage CPC + MIL semi-supervised pipeline to the binary classification of breast cancer histology images. Across five random splits, we report state-of-the-art performance with a mean validation accuracy of 95\% and an area under the ROC curve of 0.968. We further evaluate the quality of features learned via CPC relative to simple transfer learning and show that strong classification performance using CPC features can be efficiently leveraged under the MIL framework even with the feature encoder frozen.
\end{abstract}

\section{Introduction}
The standard-of-care for diagnosis and prognosis of breast cancer is the subjective interpretation of histology slides, which is both time-consuming and suffers from inter-observer variability \cite{bad}.  Deep learning has been widely applied to histology classification at the level of patches and small regions of interests (ROIs), yielding remarkable performance when sufficient labeled training data are provided \cite{bc, sup_1, sup_2, sup_3, sup_4, sup_5, sup_6, sup_7}. However, patch or pixel-level annotation is often difficult and costly to curate. By considering each labeled image as a collection of many smaller, unlabeled patches, multiple instance learning (MIL) enables training of neural networks for histopathology image classification without patch-level annotations \cite{campanella2019clinical}. An attention-based Deep MIL approach has recently been proposed which achieves state-of-the-art performance across many MIL benchmarks  \cite{ilse2018attention}.

However, direct application of deep MIL to histopathogical image analysis carries many challenges. Notably, it is common to have limited number of slides available for training, especially for rare conditions. This makes it difficult for a MIL network to adequately learn useful feature representations and as a result we found that MIL tends to drastically overfit. Another challenge is the need to process a bag of many instances at a time, usually in a single batch. This makes backpropagation infeasible due to the large size of tissue microarrays and whole slides and the memory constraints of modern GPUs. As a result, patches need to be sampled, resulting in noisy bag labels \cite{sample}, or the feature network needs to remain fixed during training to save memory. 

We propose a two-stage semi-supervised approach that attempts to help mitigate both of these key challenges by combining MIL with data-efficient self-supervised learning via contrastive predictive coding (CPC) \cite{CPC}. We demonstrate that despite limited labeled data, by taking advantage of the large amount of unlabeled instances available in a MIL dataset, we can learn rich feature representations that significantly boost downstream supervised learning performances. We also demonstrate that we can freeze the pretrained feature network and still achieve good performances under the MIL framework while saving significant amount of GPU memory required for training, which theoretically enables us to efficiently scale up to working with large weakly annotated histopathology data. To the best of our knowledge, this is the first work on weakly-supervised histology classification that relies on self-supervised feature learning using contrastive predictive coding.
\vspace{-2mm}
\section{Method}
\textbf{Deep Attention-based MIL}
Under the MIL framework, each data point is a "bag" of $N$ unordered "instances": $X_i = \{x_1, x_2, ... , x_N\}$ and a corresponding bag label $Y_i$, where $N$ can vary across bags. Each instance $x_k$ is assumed to possess an unknown instance label $y_k$, either positive or negative. The goal of MIL is to learn the bag label $Y_i$ for unseen bags, where $Y_i$ is positive if at least one instance is positive, and is negative otherwise.
Given an arbitrarily sized image and a corresponding binary label, a natural way to apply MIL is to treat each image as a bag of smaller, unlabeled patches. The current state-of-the-art deep learning-based MIL method \cite{ilse2018attention} uses a permutation invariant aggregation function called "Attention-based MIL pooling." A CNN encodes each instance $\mathbf{x_k}$ into a low-dimensional embedding $\mathbf{z_k}$. A multi-layered attention network, with parameters $\mathbf{w}, \mathbf{V}, \text{and } \mathbf{U}$ learns to assign a weight $a_k$ (Eqn 1) to each embedding and predicts a bag embedding $\mathbf{z}$ by taking their weighted average (Eqn 2). Given the aggregated features, a linear layer is finally used to predict the probability of a positive bag.
\vspace{-1mm}
\begin{equation}
a_{k}=\frac{\exp \left\{\mathbf{w}^{\top}\left(\tanh \left(\mathbf{V} \mathbf{z}_{k}^{\top}\right) \odot \operatorname{sigm}\left(\mathbf{U h}_{k}^{\top}\right)\right)\right\}}{\sum_{j=1}^{N} \exp \left\{\mathbf{w}^{\top}\left(\tanh \left(\mathbf{V} \mathbf{z}_{j}^{\top}\right) \odot \operatorname{sigm}\left(\mathbf{U z}_{j}^{\top}\right)\right)\right\}}
\end{equation}
\vspace{-0.5mm}
\begin{equation}
\mathbf{z}=\sum_{k=1}^{N} a_{k} \mathbf{z}_{k}
\end{equation}
Attention MIL pooling is trainable and allows the network to identify discriminative instances. However, when training end-to-end, the feature network might receive meaningful gradient signals from just instances that make substantial contributions to the bag representations. As a result, when labeled data are limited, due to the weak supervisory signals of MIL, the feature network might struggle to learn rich, high-level representations and the overall model can suffer from severe overfitting. To address these limitations, we propose a two-stage semi-supervised approach where we first pre-train the feature network via self-supervised feature learning that leverages information in every single instance in the dataset. During the 2nd stage of supervised learning, we use a margin-based loss function. Additionally, to prevent the attention network from overfitting by assigning high attention weights to a few negative instances, we minimize the KL-divergence between the attention weight distribution and the uniform distribution when the bag label is negative, encouraging the network to make equal usage of negative instances.

\textbf{Contrastive Predictive Coding}
A recent breakthrough in self-supervised feature learning for downstream supervised tasks is contrastive predictive coding (CPC) \cite{CPC} \cite{CPC2}. Given a data sequence $\{x_t\}$, a feature network $g_{enc}$ first encodes each observation into a low-dimensional embedding $z_t$. An auto-regressive context network $g_c$ then computes the context by aggregating all observations prior to $t$, namely, $c_t = g_c(\{z_i\}), \text{for } i \leq t$. At its core, the CPC objective aims to maximize the mutual information between the context $c_t$, the present, and future observations $z_{t+k}, k>0$. This can be efficiently achieved by using a contrastive loss. The network is tasked with correctly identifying $z_{t+k}$ among a set containing other negative samples. Prediction for $z_{t+k}$ is made linearly with weights $W_k$: $\hat{z}_{t+k}=W_{k} c_{t}$. When the bi-linear model is used to assign target probability, the CPC objective becomes the binary cross-entropy loss for the positive target: 
\begin{equation*}
\mathcal{L}_{\mathrm{CPC}}=-\underset{Z}{\mathbb{E}}\left[\log \frac{\text{exp}(z_{t+k}^T \hat{z}_{t+k})}{\sum_{z_{i} \in Z} \text{exp}(z_{i}^T \hat{z}_{t+k})}\right]
\end{equation*}


When applying to images, CPC is performed by extracting small, overlapping patches from each image in a raster scan manner to create a grid of spatially dependent data observations. The contrastive prediction task is to summarize rows of patches in the grid and then predict multiple unseen rows below from top to bottom. Patches from both the same image and other images in the mini-batch can act as negative samples and the aforementioned contrastive loss summed over each spatial location predicted \cite{CPC2}. 

Given limited labeled data, we alleviate the burden of the feature network during weakly supervised learning by pretraining the network on unlabeled instances using CPC. We train on unlabeled patches that can be readily extracted from any image in the dataset, with the hope of learning histopathology specific high-level feature representations that can be meaningfully separated among the binary classes at the instance level. 
\begin{figure}[ht]
    \centering
    \includegraphics[width=\linewidth]{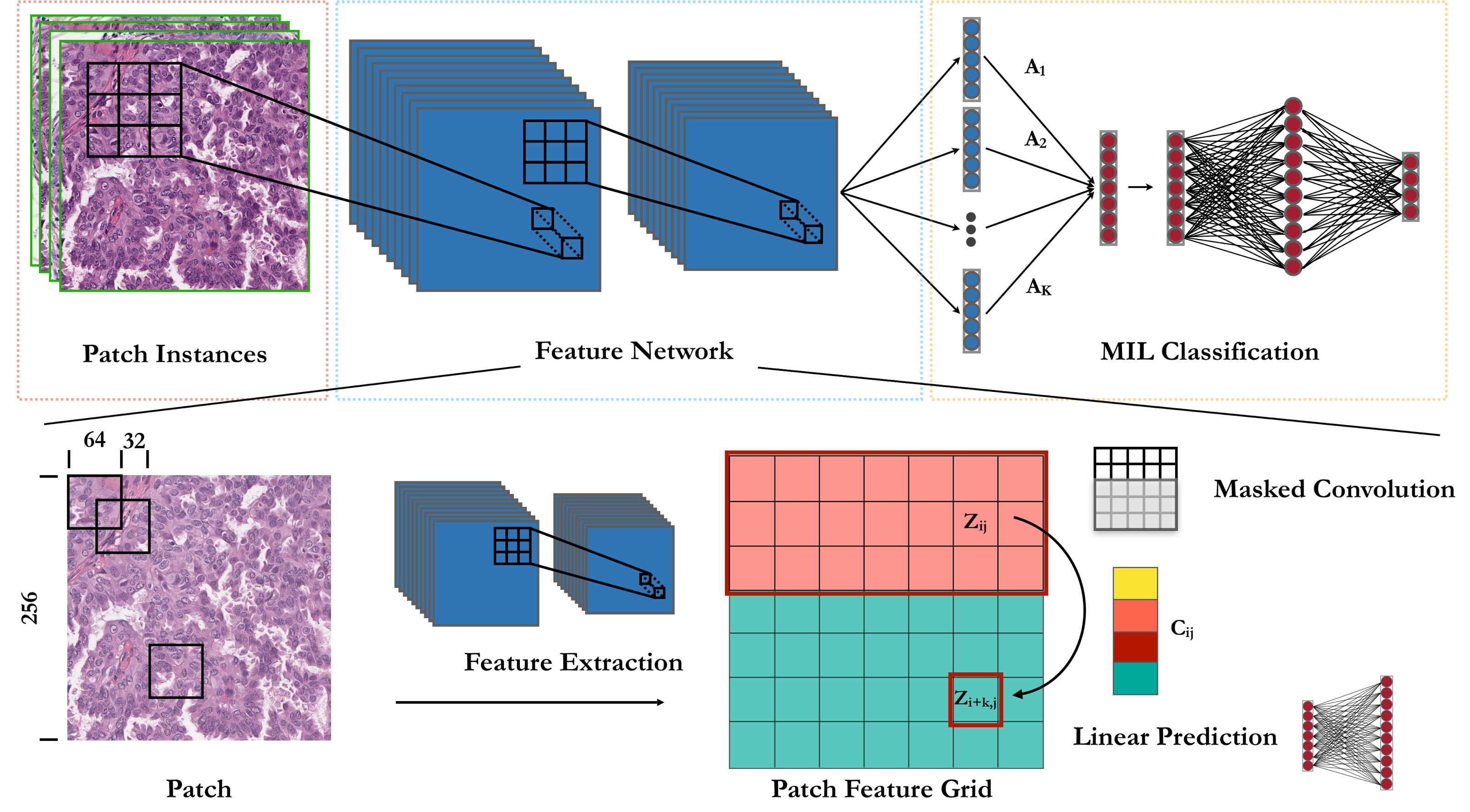}
    \caption{Schematic for proposed two-stage CPC + MIL pipeline. The Attention-based MIL network is shown at the top. The bottom shows how the feature network can be pretrained using CPC.}
\end{figure}
\vspace{-4mm}
\section{Experiments and Results}
\textbf{Dataset} 
We validate the effectiveness of CPC + MIL on the weakly supervised classification of H\&E stained breast cancer image data. The ICIAR 2018 BACH dataset \cite{bach} has a total of 400 labeled images of size $2048\times1536$. We consider the binary classification of carcinoma (in situ or Invasive) vs. non-carcinoma (Nomral or Benign). We first obtain contours for the tissue regions in each image and extract $256\times256$ patches within those regions (Appendix A).

\textbf{CPC Implementation}
We further extract small overlapping patches from each $256\times256$ patch and use a modified ResNet50 to extract a 1024-dimensional feature vector for each spatial location. The result is a grid of features with dimensions $7\times7\times1024$ for each $256\times256$ image. We use a custom, PixelCNN-style autoregressive network \cite{pixelCNN} \cite{pixelCNN2} that uses residual blocks of masked convolution layers to compute the context for each location of the feature grid. By using masked convolution, we ensure for each grid location, its receptive field covers only the rows above it (Appendix B). The result is a grid of context vectors also with dimensions $7\times7\times1024$. We use independent linear layers to predict up to 3 rows of features downstream from an arbitrarily chosen row of context vectors. 

\textbf{MIL Implementation}
During training, all patches in the same bag are collectively provided to the network in a single batch along with the image label. We observed the model is less prone to overfitting when using a smooth margin-based SVM loss \cite{smoothsvm} and our proposed regularization instead of the standard  cross-entropy loss. Using attention-MIL with no pretraining as a baseline, we evaluate the performance gained by either using transfer learning from ImageNet or self-supervised pretraining via CPC, both when the encoder is frozen and finetuned. For each split, 100 images (25\%) are randomly drawn for validation and the remaining 300 (75\%) are used for training. More training details are included in Appendix C.

\begin{table} [!h]
  \caption{Comparison between cross-entropy loss (CE) and smooth SVM loss + KL-div regularization (R) using random five-fold validation, $\pm$ standard deviation}
  \label{sample-table}
  \centering
  \begin{tabular}{lll}
    \toprule            
    Method    & Accuracy ($\%$)     & AUC ROC \\
    \toprule 
    MIL + ImageNet (CE) & 84.4 $\pm$ 9.40  &  0.933 $\pm$ 0.514 \\
    MIL + ImageNet (R)  & 86.0 $\pm$ 4.64  &  0.939 $\pm$ 0.240 \\
    \cmidrule(r){1-3}
    MIL + CPC (CE)      & 91.8 $\pm$ 7.53  &  0.959 $\pm$ 0.052 \\
    MIL + CPC (R)       & \bf 95.0 $\pm$ 2.65  & \bf 0.968 $\pm$ 0.022 \\
    \bottomrule
  \end{tabular}
  
  \caption{Comparison between different MIL pipelines using random five-fold validation, $\pm$ standard deviation (we use smooth SVM loss + KL-div regularization for all experiments in this section)}
  \centering
  \begin{tabular}{lll}
    \toprule
    Method    & Accuracy ($\%$)         & AUC ROC           \\
    \toprule 
    MIL         &   62.6 $\pm$ 11.6    & 0.611 $\pm$ 0.186     \\
    MIL + ImageNet &  86.0 $\pm$ 4.64  &  0.939 $\pm$ 0.024   \\
    MIL + CPC    & \bf 95.0 $\pm$ 2.65  & \bf  0.968 $\pm$ 0.022 \\
    \cmidrule(r){1-3}
    MIL + ImageNet, Frozen  &  82.8 $\pm$ 2.95   & 0.891 $\pm$ 0.026               \\
    MIL + CPC, Frozen       & \bf 90.6 $\pm$ 2.88   & \bf 0.939 $\pm$ 0.024            \\
    \bottomrule
  \end{tabular}
\end{table}
\begin{figure}[!h]
    \centering
    \includegraphics[width=0.75\linewidth]{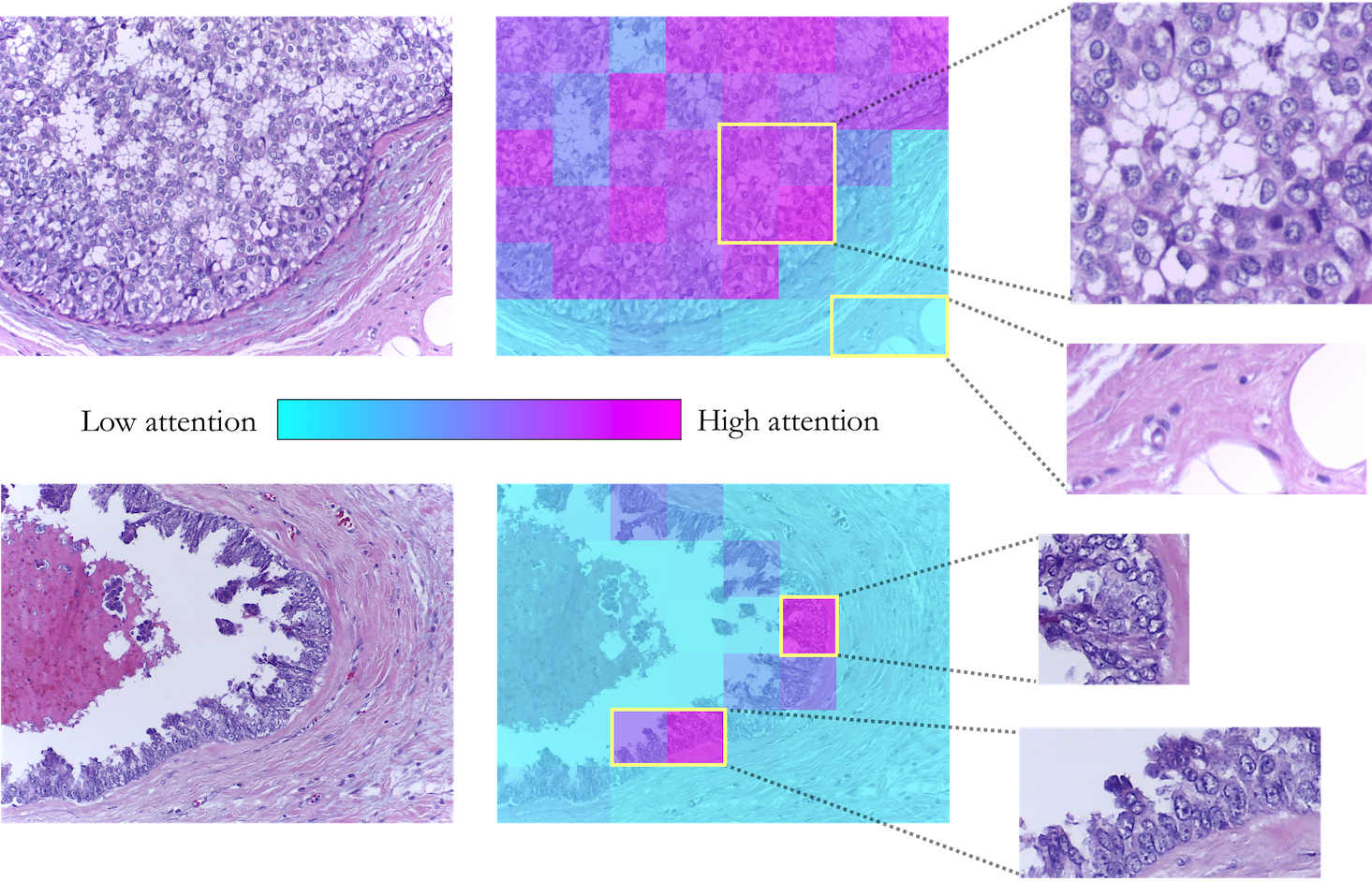}
    \caption{Attention map visualization examples for key instance identification. In general, we observe that the network mainly attends to regions of high nuclei density and masses of monomorphic cells while mostly ignoring non-informative regions including adipose tissue. However, we do notice that even when the network makes the correct classification, it sometimes fails to attend to all objects of interest including necrotic regions and suspicious, migrating cell masses that might warrant a closer inspection by a pathologist.}
    \label{fig:attention}
\end{figure}
\begin{figure}[!h]
    \centering
    \includegraphics[width=0.75\linewidth]{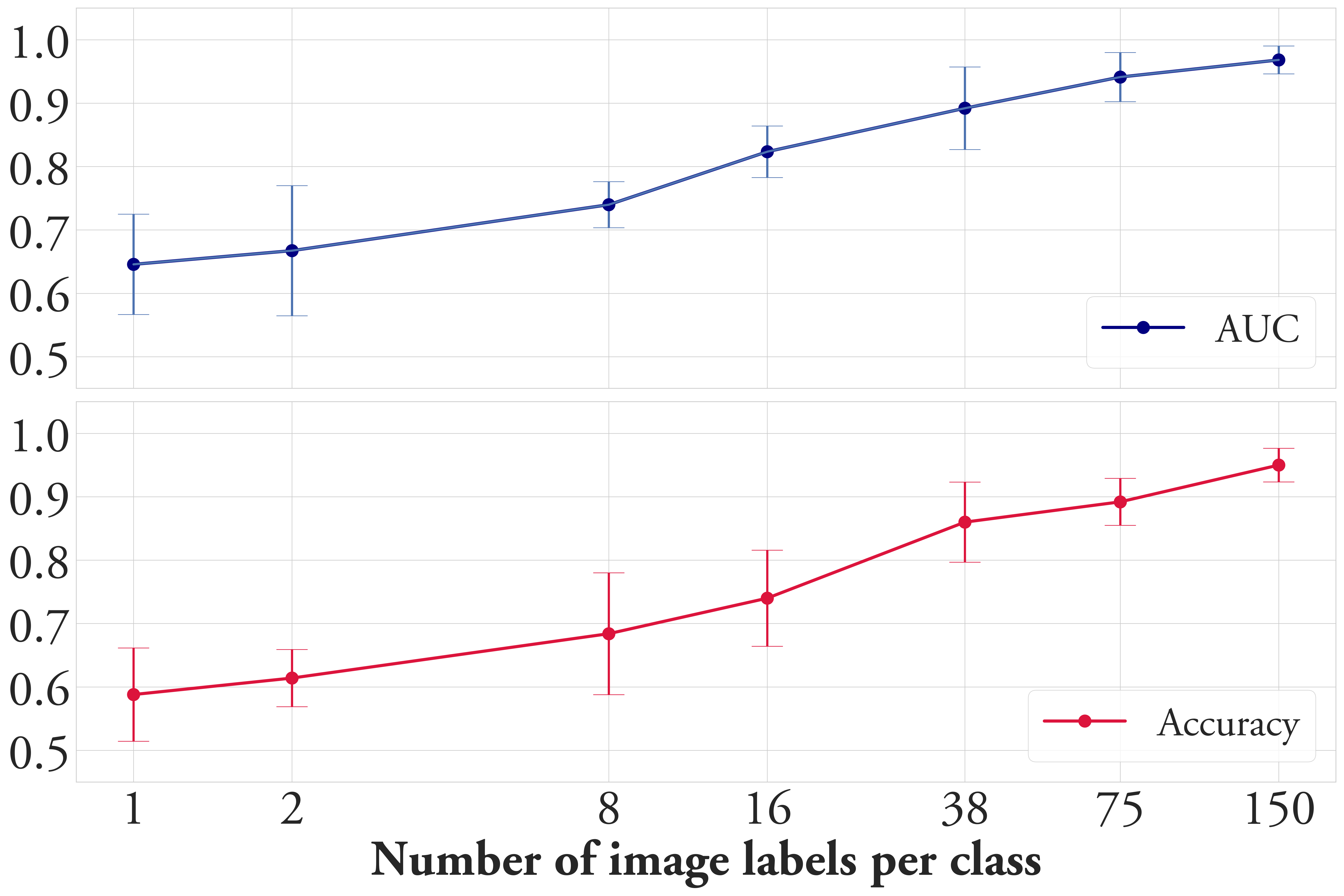}
    \caption{MIL + CPC performance for using varying numbers of labels, ranging from the minimum of 1 label per class to a maximum of 150 labels per class}
    \label{fig:eff}
\end{figure}
\vspace{-0.5mm}
We note that training CPC + MIL with smooth SVM loss + KL-div regularization appears to perform the best, outperforming MIL alone and MIL + ImageNet on every split. We also take the encoder trained on the entire dataset via CPC (unsupervised) and study the semi-supervised performance of CPC + MIL for varying numbers of labels. On the same five class-balanced random validation splits, instead of using the maximum number of 150 labels per class, training on 50\% (75 per class) of the labels still yields a mean AUC of over 0.94. Similarly, with only 16 labels per class, we can achieve a mean validation AUC of over 0.82 (Fig \ref{fig:eff}).    

Training CPC + MIL with smooth SVM loss + KL-div regularization but with the encoder completely frozen also achieves good performance. This allows us to potentially scale to larger images with much bigger bags since with the encoder frozen, the MIL network has under 800k trainable parameters. To our knowledge, this is the first MIL work to use the smooth SVM loss and KL-div regularization for negative bags. We are also the first to formally apply the MIL framework to this weakly annotated dataset and explore unsupervised feature learning to achieve state-of-the-art binary classification performance. All previously published methods such as \cite{bach_ref1}, \cite{bach_ref2} use patch-level classifiers by naively assigning the image label to every patch in the image and use transfer learning from ImageNet. Since we avoid assumptions about patch labels and can handle bags of varying sizes, our approach is flexible and potentially scalable to larger image data. 

\vspace{-1mm}
\section{Conclusions and Future Works}
We demonstrate that a deep semi-supervised approach using CPC + MIL combined with additional regularization can be effectively applied to the classification of breast cancer histology images even when MIL alone performs poorly due to overfitting on limited labeled data. Given the flexibility of our approach, we hope to scale experiments to whole slide images in the future and provide a data-efficient deep learning tool that can potentially serve as an assistive tool to reduce inter-observer variability and help pathologists improve diagnostic accuracy.

\bibliographystyle{unsrt}




\section*{Appendix A: Data Preprocessing}
\begin{figure}[ht]
    \centering
    \includegraphics[width=0.85\linewidth]{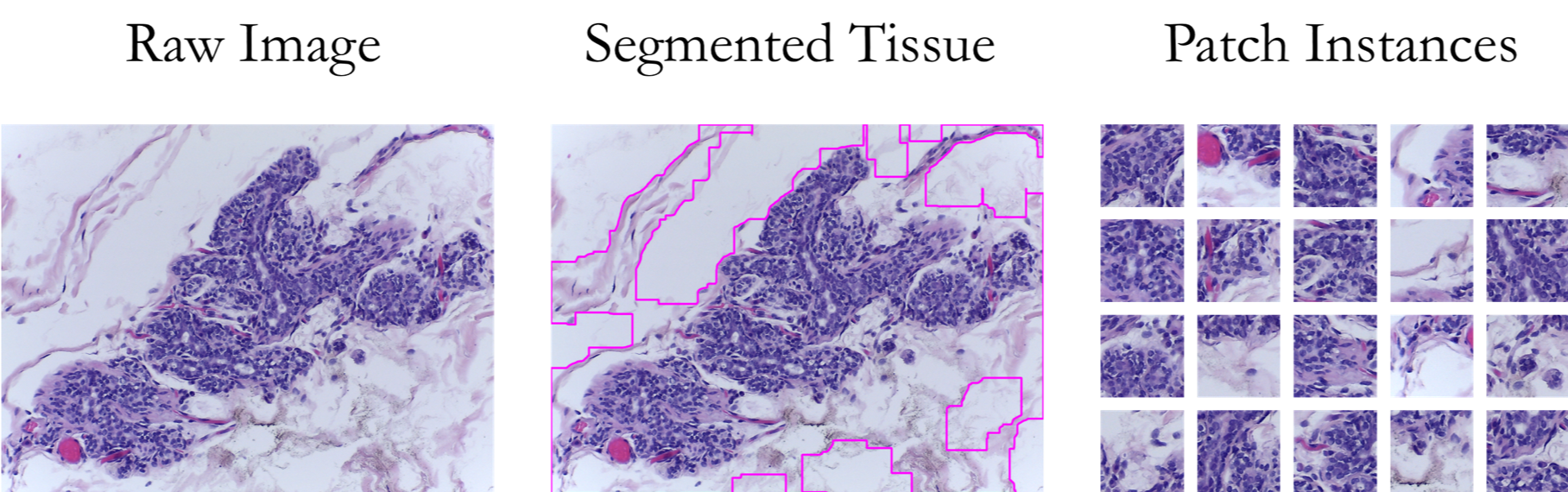}
    \caption{Data preprocessing pipeline}
    \label{fig:preprocess}
\end{figure}

Approximate tissue segmentation is performed on each raw image to obtain contours for the foreground regions (Fig \ref{fig:preprocess}).  For both CPC and MIL, only patches centered within the contours are extracted and considered.
For CPC, we extract $256\times256$ patches with 50\% overlap. For MIL, we extract patches of the same size but without overlap.

\section*{Appendix B: CPC Implementation}
We further extract small, $64\times64$ patches with 50\% overlap from each bigger $256\times256$ unlabeled patch. During training, we apply flips to both the large and small patches as data augmentation. At the small patch level, we also adopt the color channel dropping regularization and spatial jittering \cite{CPC2}. For the feature encoder, we use a ResNet50 without Batchnorm layers. We truncate the network after the 3rd residual block and spatial pool to compute a 1024-dimensional vector for each $64\times64$ patch. The context network contains a series of masked-convolution blocks (Table 3). Masks are applied to outputs of convolutions such that everything below the current pixel is made hidden. 

\begin{figure}[h]
    \begin{subfigure}{0.45\textwidth}
        \centering
        \includegraphics[width=0.75\linewidth]{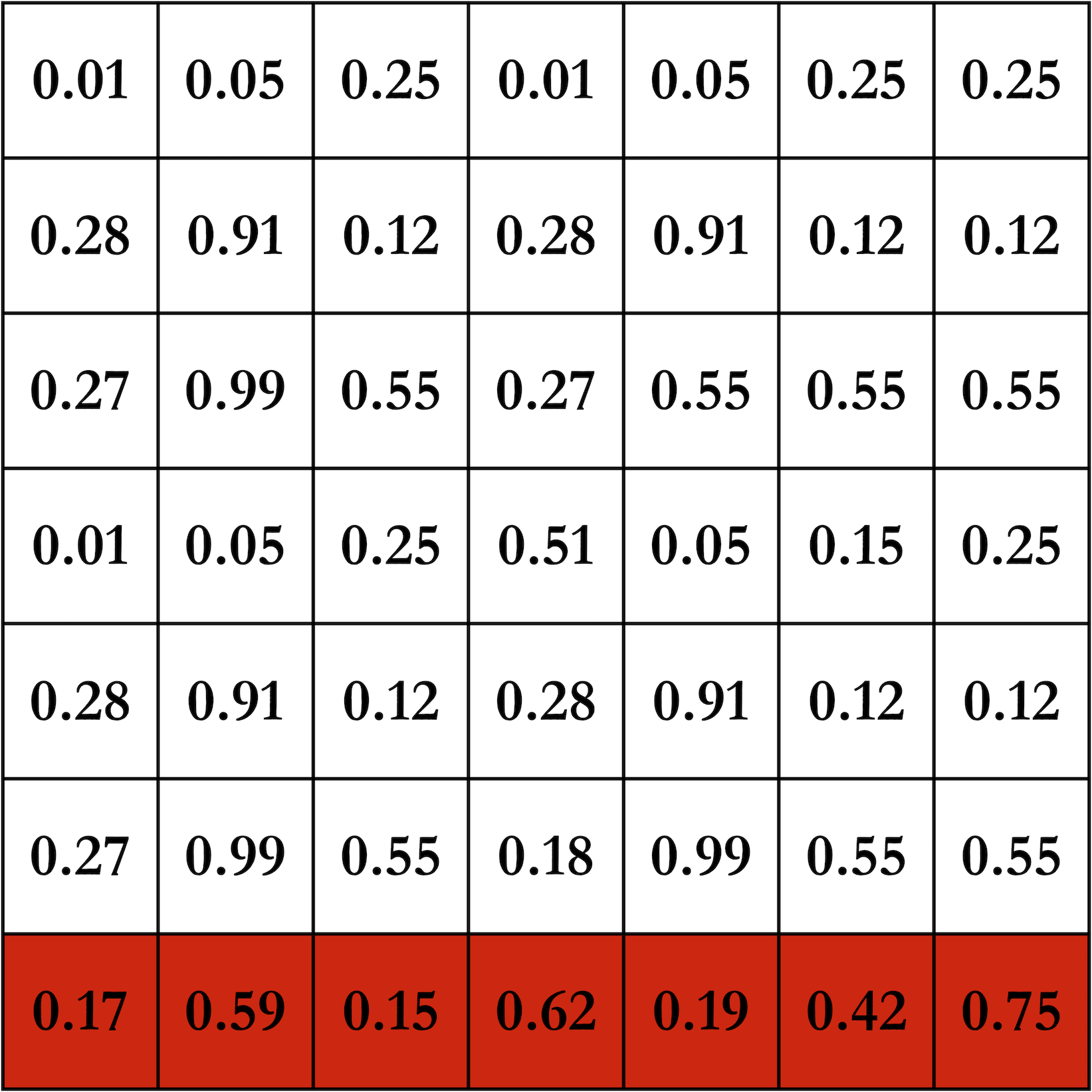}
        \caption{Original feature map}
    \end{subfigure}
    \begin{subfigure}{0.45\textwidth}
        \centering
        \includegraphics[width=0.75\linewidth]{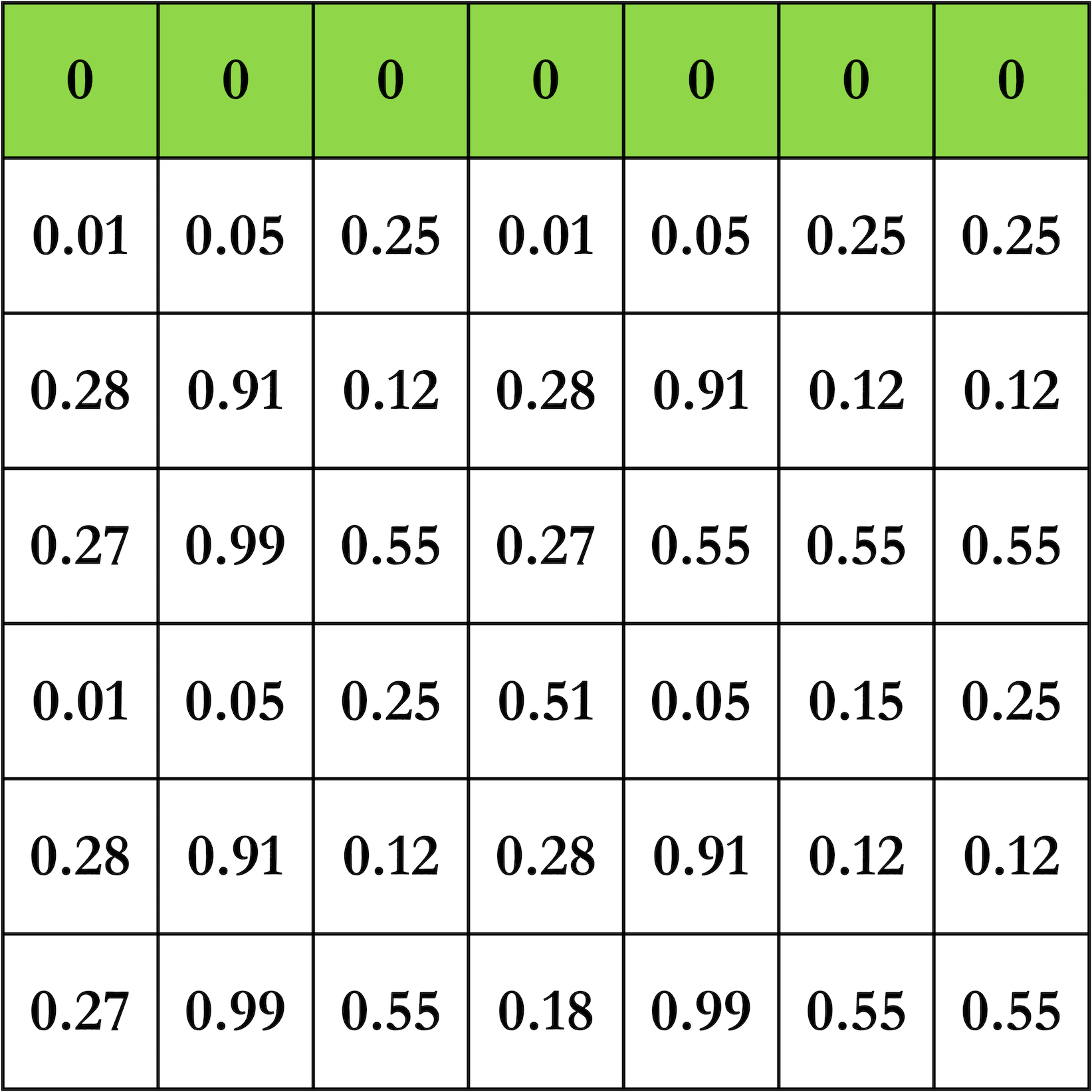}
        \caption{Downshifted feature map}
    \end{subfigure}
    
    \caption{Pad and downshift example on $7\times7$ feature map}
    \label{fig:pad_downshift}
\end{figure}

\textbf{Masked Conv Block A} pads the top of the feature grid with zeros and removes the bottom row. Together with convolution masks, this "pad and downshift" operation (Fig \ref{fig:pad_downshift}) ensures that the receptive field of each output neuron only sees the rows that lie above it in the original feature map while avoiding any receptive field blind spots associated with the original PixelCNN implementation \cite{pixelCNN}. A $7\times7$ masked convolution followed by Batchnorm is then applied and reduces the number of features from 1024 to 256. The feature maps are then split in half. A gated activation function \cite{pixelCNN2} applies a tanh non-linearity to half of the feature maps and a sigmoid function to the other half and element-wise multiply the two halves to compute the final activation. 

\textbf{Masked Conv Block B} applies a $3\times3$ masked convolution that increases the number of feature maps from 128 to 256, followed by Bachnorm and the same gated activation that halves the number of channels and add the original input to the output (residual connection). The spatial resolution and depth of the feature map after each Masked Conv Block B therefore remains constant. The last block of the network uses $1\times1$ convolution to increase the feature dimension from 128 back to 1024. 

\begin{table}[!h]
  \caption{Context network architecture}
  \centering
  \begin{tabular}{lll}
    \toprule
    Layer    & Type    \\
    \cmidrule(r){1-2}
    1 &      Masked Conv Block A    \\
    2 &      Masked Conv Block B     \\
     &      ... \\
    11 &     Masked Conv Block B \\
    12 &     ReLU + Conv $1\times1$ + Batchnorm2d + ReLU \\
    \bottomrule
  \end{tabular}
\end{table}
We predict up to 3 unseen rows of features for each image, using a batch size of 16 per GPU on 4 NVIDIA Tesla P100's and a learning rate of 1e-3 with Adam. Our code is implemented in Pytorch and will be made available in the future. 

\section*{Appendix C: MIL Implementation}
We keep the same ResNet encoder architecture as in CPC. However, when we do transfer learning from ImageNet, we add the Batchnorm layers back to the ResNet to be consistent with the original settings. 
When the encoder is finetuned end-to-end, we connect a compact gated attention-MIL network \cite{ilse2018attention} with 256 hidden units directly to the encoder. It predicts an attention score for each 1024-dimensional instance embedding in the bag and their weighted average is efficiently computed by using matrix multiplication.

When training with a frozen encoder, we add an additional trainable fully-connected layer to reduce the dimensions of CPC features from 1024 to 512 before connecting to the attention network. The bag representations are therefore of length 512 instead of 1024. We found that the additional linear layer greatly improve training speed and performance. This likely signals that the binary categories are not completely linearly separable in the CPC feature space and would benefit from additional transformations before classification.

For all scenarios, we utilize light Dropout in the attention network as well as flips, color-jittering  \cite{colorjitter} and spatial jittering as data augmentation during training. For each split, 100 random images (25\%) are drawn for validation and the remaining 300 (75\%) are used for training. Train/validation splits are class-balanced and are held consistent across experiments by using fixed seeding. We train using an effective batch size of 4 on multiple NVIDIA Tesla P100 or K80 GPUs for up to 100 epochs. We use Adam with a learning rate of 5e-5 when finetuning and 2e-4 when the encoder is frozen. The earlystopping criterion is set to when the validation loss does not improve for more than 25 epochs. 

\end{document}